\documentclass[10pt,twocolumn,letterpaper]{article}

\usepackage{cvpr}
\usepackage{times}
\usepackage{epsfig}
\usepackage{graphicx}
\usepackage{amsmath}
\usepackage{amssymb}
\usepackage{tabularx}
\usepackage{multirow}
\usepackage{amssymb}
\usepackage{amsmath}
\usepackage{pifont}
\usepackage[FIGTOPCAP]{subfigure}
\usepackage[super]{nth}
\newcommand{\cmark}{\ding{51}}%
\newcommand{\xmark}{\ding{55}}%
\usepackage{color}
\usepackage{balance}
\usepackage{booktabs}
\usepackage{overpic}
\newcommand\mypara[1]{\vspace{1mm}\noindent\textbf{#1}}

\usepackage[breaklinks=true,bookmarks=false]{hyperref}

\cvprfinalcopy 

\def\cvprPaperID{236} 

\begin{document}

\title{SelFlow: Self-Supervised Learning of Optical Flow}

\author{Pengpeng Liu$^\dag$\thanks{Work mainly done during an internship at Tencent AI Lab.}, Michael Lyu$^\dag$,
Irwin King$^\dag$,  Jia Xu$^\S$ \\
$^\dag$  The Chinese University of Hong Kong,
$^\S$ Tencent AI Lab
}

\maketitle

\begin{abstract}
We present a self-supervised learning approach for optical flow. Our method distills reliable flow estimations from non-occluded pixels, and uses these predictions as ground truth to learn optical flow for hallucinated occlusions. We further design a simple CNN to utilize temporal information from multiple frames for better flow estimation. These two principles lead to an approach that yields the best performance for unsupervised optical flow learning on the challenging benchmarks including MPI Sintel, KITTI 2012 and  2015. More notably, our self-supervised pre-trained model provides an excellent initialization for supervised fine-tuning. Our fine-tuned models achieve state-of-the-art results on  all three datasets. At the time of writing, we achieve EPE=4.26 on the Sintel benchmark, outperforming all submitted methods.
\end{abstract}

\section{Introduction}
Optical flow estimation is a core building block for a variety of computer vision systems~\cite{menze2015object,chauhan2013moving,simonyan2014two,Bonneelsiggraph2015}. Despite decades of development, accurate flow estimation remains an open problem due to one key challenge: occlusion.
Traditional approaches minimize an energy function to encourage association of visually similar pixels and  regularize incoherent motion to propagate flow estimation from non-occluded pixels to occluded pixels~\cite{horn1981determining,brox2004high,brox2011large,revaud2015epicflow}. However, this family of methods is often time-consuming and not applicable for real-time applications.

\begin{figure*}[t]
 \centering
   \includegraphics[width=0.95\textwidth]{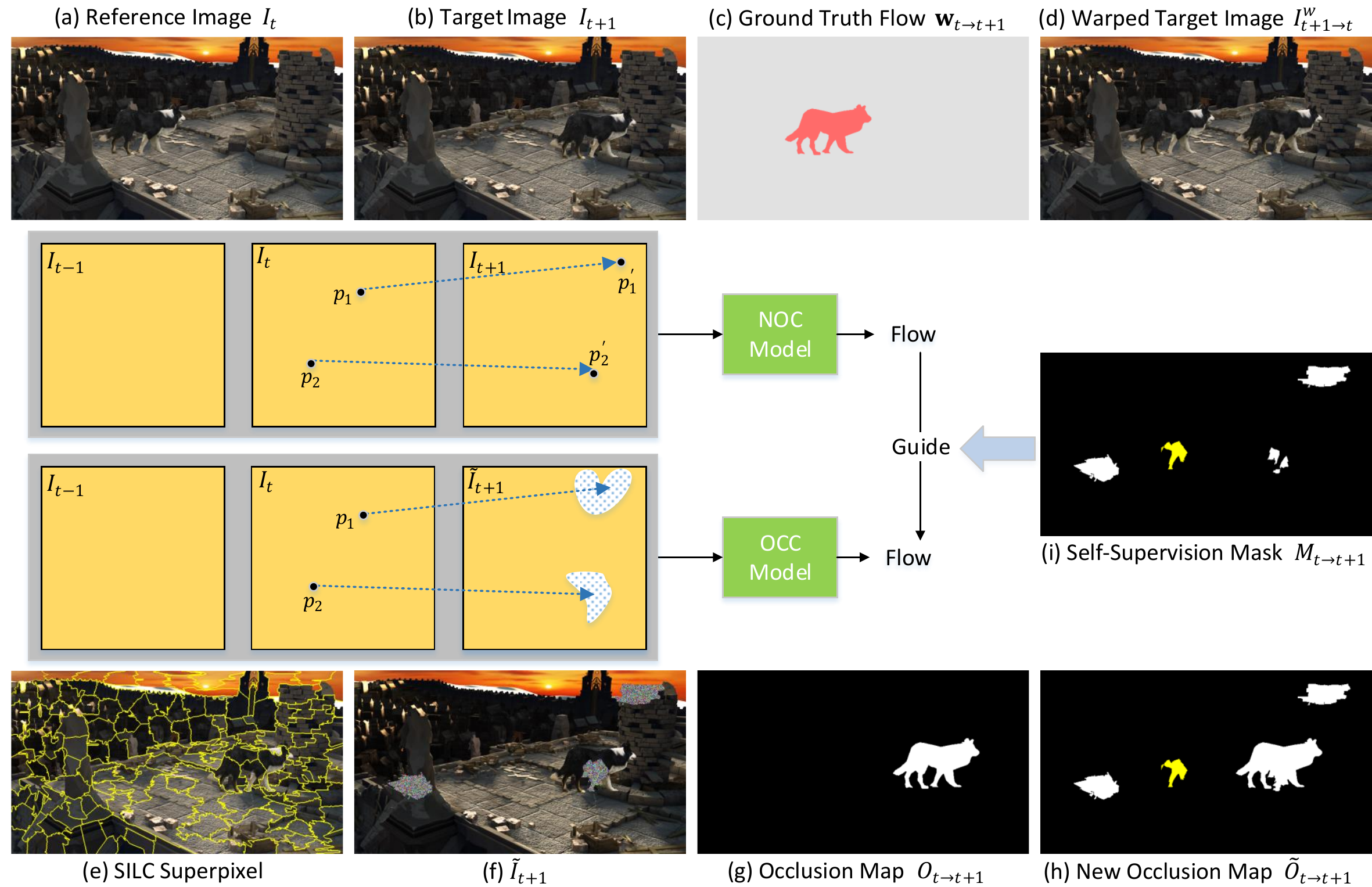}
 \caption{A toy example to illustrate our self-supervised learning idea.
We first train our NOC-model with the classical photometric loss (measuring the difference between the reference image (a) and
the warped target image(d)), guided by the occlusion map (g). Then we perturbate randomly selected superpixels in the target image (b) to hallucinate occlusions. Finally,  we use reliable flow estimations from our NOC-Model to guide the learning of our OCC-Model for those newly occluded pixels (denoted by self-supervision mask (i), where value 1 means the pixel is non-occluded in (g) but occluded in (h)). Note the yellow region is part of the moving dog. Our self-supervised approach  learns optical flow for both moving objects and static scenes.
}
 \label{Framework}
 \end{figure*}

Recent studies learn to estimate optical flow end-to-end from images using convolutional neural networks (CNNs)~\cite{dosovitskiy2015flownet,ranjan2017optical,ilg2017flownet,hui18liteflownet,sun2018pwc}.
However, training fully supervised CNNs requires a large amount of labeled training data, which is extremely difficult to obtain for optical flow, especially when there are occlusions.
Considering the recent performance improvements obtained when employing hundreds of millions of labeled images~\cite{sun2017revisiting}, it is obvious that the size of  training data is a key  bottleneck for optical flow estimation.

In the absence of large-scale real-world annotations, existing methods turn to pre-train on synthetic labeled datasets \cite{dosovitskiy2015flownet,mayer2016large} and then fine-tune on small annotated datasets~\cite{ilg2017flownet,hui18liteflownet,sun2018pwc}. However, there usually exists a large gap between the distribution of synthetic data and natural scenes. In order to train a stable model,  we have to carefully follow specific learning schedules across different datasets~\cite{ilg2017flownet,hui18liteflownet,sun2018pwc}.

One promising direction is to develop unsupervised optical flow learning methods that benefit from unlabeled data. The basic idea is to warp the target image towards the reference image according to the estimated optical flow, then minimize the difference between the reference image and the warped target image using a photometric loss~\cite{jason2016back,ren2017unsupervised}.
Such idea works well for non-occluded pixels but turns to provide misleading information for occluded pixels. Recent  methods propose to exclude those occluded pixels when computing the photometric loss or employ additional spatial and temporal smoothness terms to regularize flow estimation~\cite{Meister:2018:UUL,wang2018occlusion,Janai2018ECCV}. Most recently, DDFlow~\cite{Liu:2019:DDFlow} proposes a data distillation approach, which employs random cropping  to create occlusions for self-supervision. Unfortunately, these methods fails to generalize well for all natural occlusions. As a result, there is still a large performance gap comparing unsupervised methods with state-of-the-art fully supervised methods.

Is it possible to effectively learn optical flow with occlusions? In this paper, we  show that a self-supervised  approach can learn to estimate optical flow with any form of occlusions from unlabeled data. Our work is based on distilling reliable flow estimations from non-occluded pixels, and using these predictions to guide the optical flow learning for hallucinated occlusions. Figure \ref{Framework} illustrates our idea to create synthetic occlusions by perturbing superpixels. We further utilize temporal information from multiple frames to improve flow prediction accuracy within a simple CNN architecture.
The resulted learning approach yields the highest accuracy among all unsupervised optical flow learning methods on Sintel and KITTI benchmarks.

Surprisingly, our self-supervised pre-trained model  provides an excellent initialization for supervised fine-tuning. At the time of writing, our fine-tuned model achieves the highest reported accuracy (EPE=4.26) on the Sintel benchmark. Our approach also significantly outperforms all published  optical flow
methods on the KITTI 2012 benchmark, and achieves highly competitive results on the KITTI 2015 benchmark. To the best of our knowledge,  it is the first time that a supervised learning method achieves such  remarkable accuracies  without using  any external labeled data.

\section{Related Work}
\mypara{Classical   Optical Flow Estimation.}
Classical variational approaches model optical flow estimation as an energy minimization problem based on brightness constancy and spatial smoothness~\cite{horn1981determining}. Such methods are effective for small motion, but  tend to fail when displacements are large. Later works integrate feature matching to initialize sparse matching, and then interpolate into dense flow maps  in a pyramidal coarse-to-fine manner~\cite{brox2011large,weinzaepfel2013deepflow,revaud2015epicflow}. Recent works use convolutional neural networks (CNNs) to improve sparse matching by  learning  an effective  feature embedding~\cite{XRK2017,bailer2017cnn}. However,  these methods are often computationally expensive and can not be trained  end-to-end. One natural  extension to improve  robustness and accuracy for flow estimation is to incorporate temporal information over multiple frames. A straightforward way is to add temporal constraints such as constant velocity~\cite{janai2017slow,kennedy2015optical,sun2010layered}, constant acceleration~\cite{volz2011modeling,black1991robust}, low-dimensional linear subspace~\cite{irani1999multi}, or rigid/non-rigid segmentation~\cite{wulff2017optical}. While these formulations are  elegant and well-motivated, our method is much simpler and does not rely on any assumption of the data. Instead, our approach directly learns optical flow for a much wider range of challenging cases existing in the data.

\begin{figure}[t]
 \centering
   \includegraphics[width=0.48\textwidth]{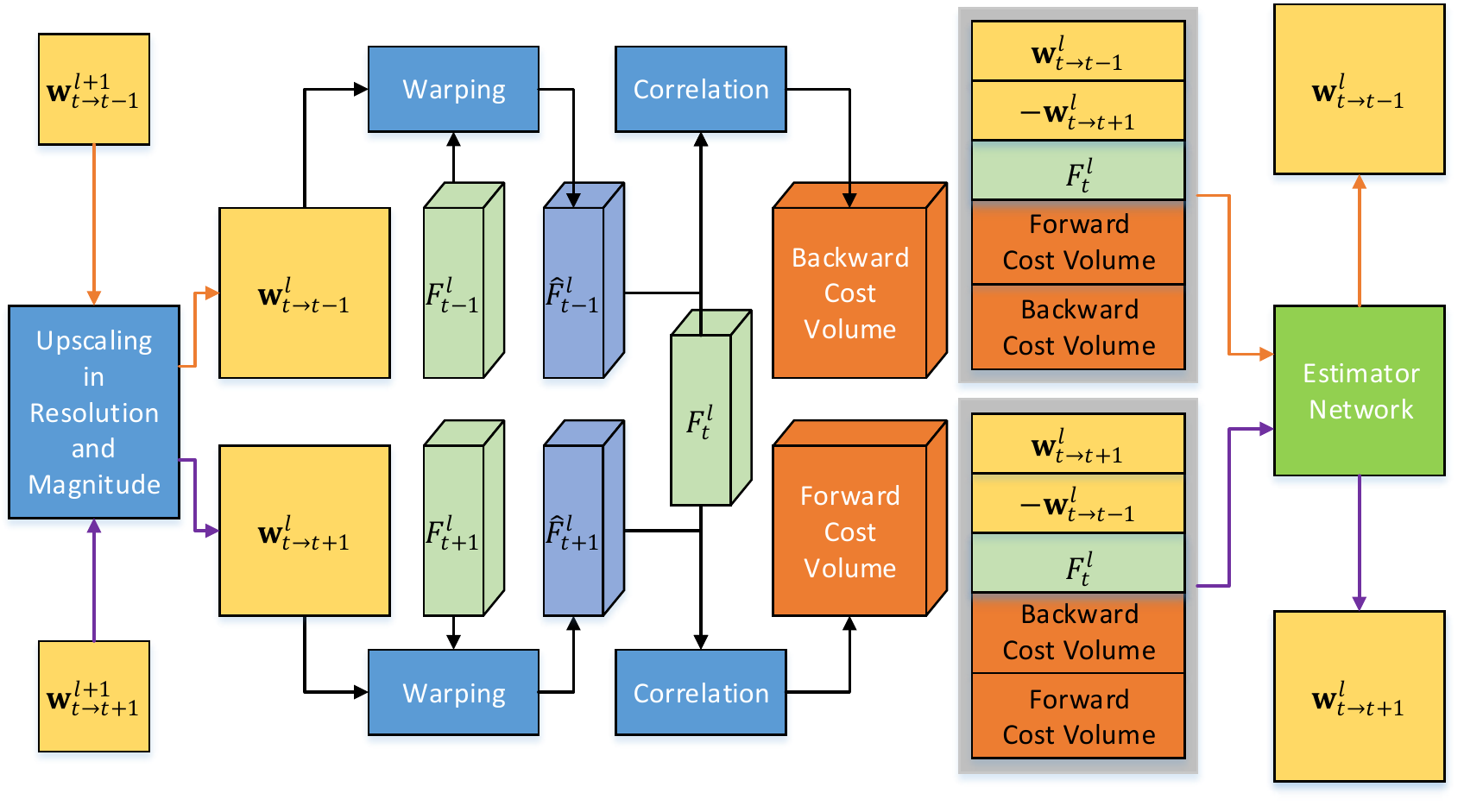}
 \caption{Our network architecture at each level (similar to PWC-Net~\cite{sun2018pwc}). $\dot{\textbf{w}}^l$ denotes the initial coarse flow of level $l$ and $\hat{F}^l$ denotes the warped feature representation. At each level, we swap the initial flow and cost volume as input to estimate both forward and backward flow concurrently. Then these estimations are passed to layer $l-1$ to estimate higher-resolution flow.
}
 \label{NetworkArchitecture}
 \end{figure}

\mypara{Supervised Learning of Optical Flow.}
One promising direction is to learn optical flow with CNNs. FlowNet~\cite{dosovitskiy2015flownet} is the first  end-to-end optical flow learning framework. It takes two consecutive images as input and outputs a dense  flow map. The following work FlowNet 2.0~\cite{ilg2017flownet} stacks several basic FlowNet models for iterative refinement, and significantly improves the accuracy. SpyNet~\cite{ranjan2017optical} proposes to warp images at multiple scales to cope with large displacements, resulting in a compact spatial pyramid network. Recently, PWC-Net~\cite{sun2018pwc} and LiteFlowNet~\cite{hui18liteflownet} propose to warp features extracted from CNNs  and achieve state-of-the-art results with lightweight framework. However, obtaining high accuracy with these CNNs requires pre-training on multiple synthetic datasets and follows specific training schedules~\cite{dosovitskiy2015flownet,mayer2016large}. In this paper, we reduce  the reliance on pre-training with synthetic data, and propose an effective self-supervised training method with unlabeled data.

\mypara{Unsupervised Learning of Optical Flow.}
Another interesting line of work is unsupervised optical flow learning. The basic principles are based on brightness constancy and spatial smoothness~\cite{jason2016back,ren2017unsupervised}. This leads to the most popular photometric loss, which measures the difference between the reference image and the warped image. Unfortunately, this loss does not hold for occluded pixels. Recent studies propose to first obtain an occlusion map and then exclude those occluded pixels when computing the photometric difference~\cite{Meister:2018:UUL,wang2018occlusion}. Janai~\etal~\cite{Janai2018ECCV} introduces   to estimate optical flow with a multi-frame formulation and more advanced occlusion reasoning, achieving state-of-the-art unsupervised results. Very recently, DDFlow~\cite{Liu:2019:DDFlow} proposes a data distillation approach to learning the optical flow of occluded pixels, which works particularly well for pixels near  image boundaries.  Nonetheless, all these unsupervised learning methods  only handle specific cases of occluded pixels. They lack the ability to reason about the optical flow of all possible occluded pixels. In this work, we address this issue by a superpixel-based  occlusion hallucination technique.

\mypara{Self-Supervised Learning.}
Our work is closely related to the family of self-supervised learning methods,  where the supervision signal is purely generated from the data itself.
It is widely used for learning  feature representations from unlabeled data~\cite{jing2019self}. A pretext task is usually employed, such as image inpainting~\cite{pathak2016context}, image colorization~\cite{larsson2017colorization}, solving Jigsaw puzzles~\cite{noroozi2016unsupervised}. Pathak~\etal~\cite{pathak2017learning} propose to explore low-level motion-based cues to learn feature representations without manual supervision. Doersch~\etal~\cite{doersch2017multi} combine multiple self-supervised learning tasks to train a single visual representation. In this paper, we make  use of the domain knowledge of optical flow, and take reliable predictions of non-occluded pixels as the self-supervision signal to guide our optical flow learning of occluded pixels.

\section{Method}
In this section, we present our  self-supervised approach to learning  optical flow from unlabeled data. To this end, we train two CNNs (NOC-Model and OCC-Model)  with the same network architecture. The former  focuses on accurate flow estimation for  non-occluded pixels, and the latter learns to predict  optical flow for all pixels. We distill  reliable non-occluded flow estimations from NOC-Model to guide the learning of OCC-Model for those occluded pixels. Only OCC-Model is needed at testing. We build our network based on PWC-Net~\cite{sun2018pwc} and further  extend it to multi-frame optical flow estimation (Figure~\ref{NetworkArchitecture}). Before describing our approach in detail, we first  define our notations.

\begin{figure}[t]
 \centering
 \includegraphics[width=0.48\textwidth]{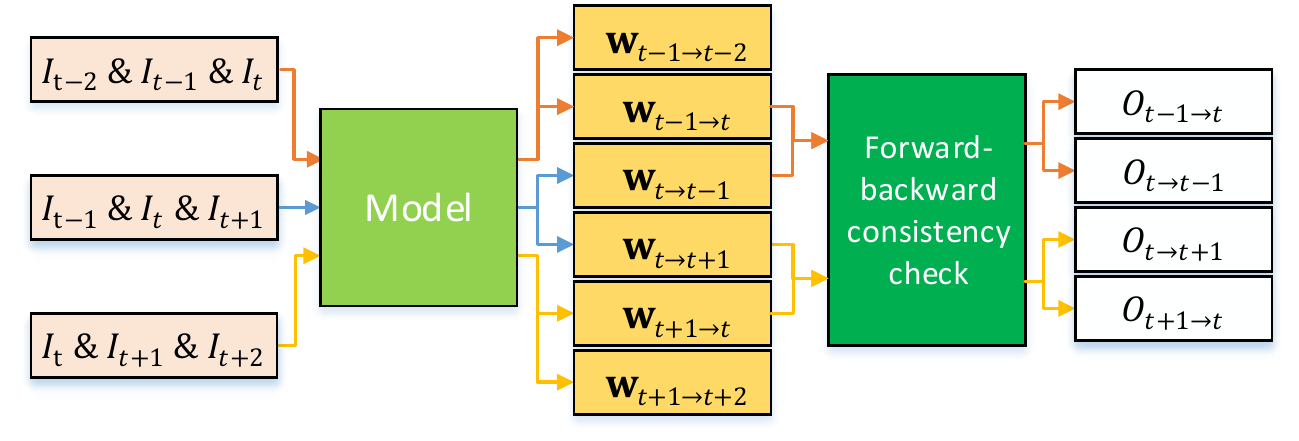}
 \caption{Data flow for self-training with multiple-frame.
 To estimate occlusion map for three-frame flow learning, we use five images as input. This way, we can conduct a forward-backward consistency check to estimate  occlusion maps between $I_t$ and $I_{t+1}$, between $I_t$ and $I_{t-1}$ respectively.
}
 \label{OcclusionEstimation}
 \end{figure}

 \begin{figure*}[t]
 \centering
   \includegraphics[width=1\textwidth]{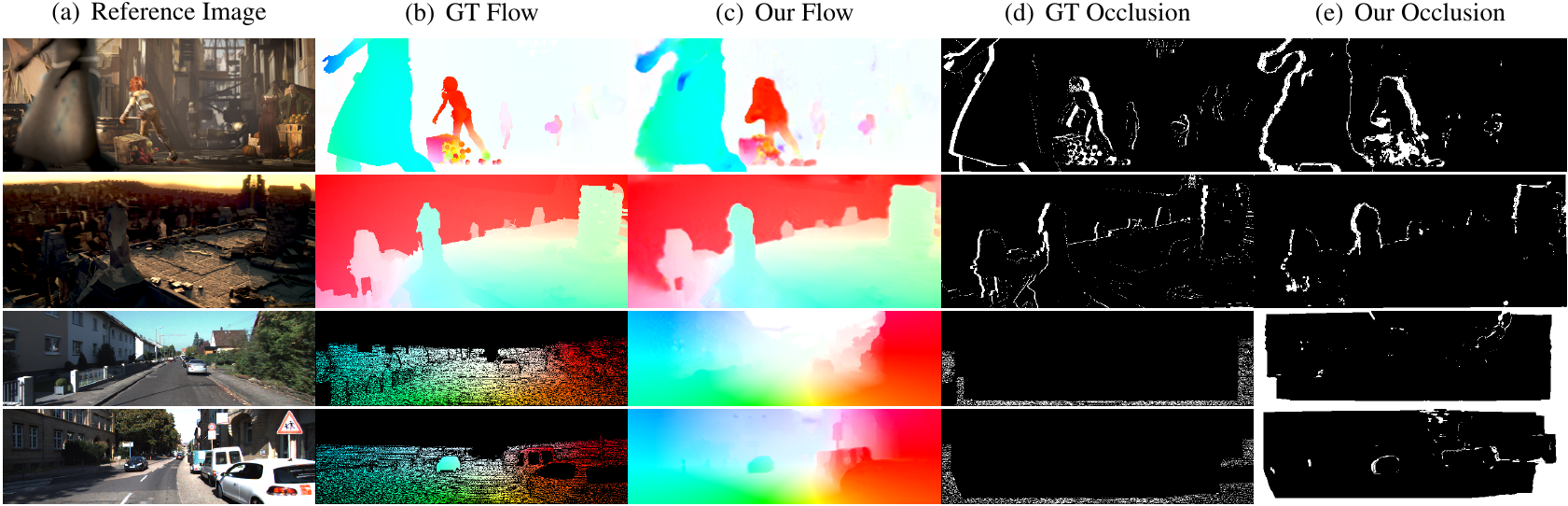}
 \caption{Sample unsupervised results on Sintel and KITTI dataset. From top to bottom, we show  samples from  Sintel Final, KITTI 2012 and KITTI 2015. Our model can estimate both accurate flow and occlusion map. Note that on KITTI datasets, the occlusion maps are sparse, which only contain pixels moving out of the image boundary.
}
\label{QualitativeResult}
\end{figure*}

\subsection{Notation}
Given three consecutive RGB images $I_{t-1}$, $I_{t}$, $I_{t+1}$, our goal is to estimate the forward optical flow from $I_t$ to $I_{t+1}$. Let $\textbf{w}_{i \to j}$ denote the flow from $I_i$ to $I_j$, e.g., $\textbf{w}_{t \to t+1}$ denotes the forward flow from $I_t$ to $I_{t+1}$,  $\textbf{w}_{t \to t-1}$ denotes the backward flow from $I_t$ to $I_{t-1}$. After obtaining optical flow, we can backward warp the target image to reconstruct the reference image using Spatial Transformer Network~\cite{jaderberg2015spatial,wang2018occlusion}. Here, we use $I_{j \to i}^w$ to denote warping $I_j$ to $I_i$ with flow $\textbf{w}_{i \to j}$. Similarly, we use $O_{i \to j}$ to denote the occlusion map from $I_i$ to $I_j$, where value 1 means the pixel in $I_i$ is not visible in $I_j$.

In our self-supervised setting, we create the new target image $\widetilde{I}_{t+1}$ by injecting
random noise on superpixels for occlusion generation. We can inject noise to any of three consecutive frames and even multiple of them as shown in Figure~\ref{Framework}. For brevity, here we choose $I_{t+1}$ as an example. If we let $I_{t-1}$, $I_t$ and  $\widetilde{I}_{t+1}$ as input, then $\widetilde{\textbf{w}}, \widetilde{O}, \widetilde{I}^w$ represent the generated optical flow, occlusion map and warped image respectively.

\subsection{CNNs for Multi-Frame Flow Estimation}
In principle, our method can utilize any CNNs. In our implementation, we build on top of the seminar PWC-Net~\cite{sun2018pwc}.
PWC-Net employs pyramidal processing to increase the flow resolution in a coarse-to-fine manner and utilizes feature warping, cost volume construction to estimate optical flow at each level. Based on these principles, it has achieved state-of-the-art performance with a compact model size.

As shown in Figure~\ref{NetworkArchitecture}, our three-frame flow estimation network structure is built upon two-frame PWC-Net with several modifications to aggregate temporal information. First, our network takes three images as input, thus produces three feature representations $F_{t-1}$, $F_t$ and $F_{t+1}$. Second, apart from forward flow $\textbf{w}_{t \to t+1}$ and forward cost volume, out model also computes backward flow $\textbf{w}_{t \to t-1}$ and backward cost volume at each level simultaneously. Note that when estimating forward flow, we also utilize the initial backward flow and backward cost volume information. This is because  past frame $I_{t-1}$ can provide very valuable information, especially for those regions that are occluded in the future frame $I_{t+1}$ but not occluded in $I_{t-1}$. Our network combines all this information together and therefore estimates optical flow more accurately. Third, we stack initial forward flow $\dot{\textbf{w}}_{t \to t+1}^l$,
minus initial backward flow $-\dot{\textbf{w}}_{t+1 \to t}^l$, feature of reference image $F_t^l$, forward cost volume and backward cost volume to estimate the forward flow at each level. For backward flow, we just swap the flow and cost volume as input. Forward and backward flow estimation networks share the same network structure and weights. For initial flow at each level, we upscale optical flow of the next level both in resolution and magnitude.

\begin{table*}[t]
\centering
\resizebox{\textwidth}{!}{
\begin{tabular}{ c  l  c  c  c  c  c  c  c   c  c}
 \toprule
   &   \multirow{2}{*}{Method} & \multicolumn{2}{c}{Sintel Clean} & \multicolumn{2}{c}{Sintel Final} & \multicolumn{3}{c}{KITTI 2012} & \multicolumn{2}{c}{KITTI 2015} \\
   \cmidrule(l{3mm}r{3mm}){3-4}    \cmidrule(l{3mm}r{3mm}){5-6} \cmidrule(l{3mm}r{3mm}){7-9}    \cmidrule(l{3mm}r{3mm}){10-11}
                   &  & train & test & train & test & train & test & test(Fl) & train & test(Fl) \\
  \midrule
  \multirow{7}{*}{\rotatebox[origin=c]{90}{Unsupervised}}
   & BackToBasic+ft~\cite{jason2016back}       & --        & --        & --        & --        & 11.3      & 9.9   & --& --     & --      \\
   & DSTFlow+ft~\cite{ren2017unsupervised}     & (6.16)    & 10.41     & (6.81)    & 11.27     & 10.43     & 12.4  & --& 16.79  & 39\%    \\
   & UnFlow-CSS~\cite{Meister:2018:UUL}     & --        & --        & (7.91)    & 10.22     & 3.29      & --    &  --& 8.10   & 23.30\% \\
   & OccAwareFlow+ft~\cite{wang2018occlusion}  & (4.03)    & 7.95      & (5.95)    & 9.15      & 3.55      & 4.2     &  --& 8.88   & 31.2\%      \\
   & MultiFrameOccFlow-None+ft~\cite{Janai2018ECCV}& (6.05)    & --        & (7.09)    & --        & --        & --  &  --& 6.65   & --      \\
   & MultiFrameOccFlow-Soft+ft~\cite{Janai2018ECCV}& (3.89)    & 7.23      & (5.52)    & 8.81      & --        & --  &  --&6.59   & 22.94\% \\
   & DDFlow+ft~\cite{Liu:2019:DDFlow}          &(2.92) & \textbf{6.18} & 3.98  & 7.40 & 2.35 & 3.0 & 8.86\% & 5.72 & 14.29\% \\
   &Ours & \textbf{(2.88)} & 6.56  &\textbf{(3.87)}& \textbf{6.57} &\textbf{1.69}&\textbf{2.2}&\textbf{7.68\%}&\textbf{4.84}&\textbf{14.19\%}\\
   \midrule
   \multirow{15}{*}{\rotatebox[origin=c]{90}{Supervised}}
   & FlowNetS+ft~\cite{dosovitskiy2015flownet} & (3.66)    & 6.96      & (4.44)    & 7.76      & 7.52      & 9.1   &44.49\%& -- & --      \\
   & FlowNetC+ft~\cite{dosovitskiy2015flownet} & (3.78)    & 6.85      & (5.28)    & 8.51      & 8.79      & -- & -- & -- & -- \\
   & SpyNet+ft~\cite{ranjan2017optical}        & (3.17)    & 6.64      & (4.32)    & 8.36      & 8.25      & 10.1  & 20.97\% & --     & 35.07\%  \\
   & FlowFieldsCNN+ft~\cite{bailer2017cnn}        & --        & 3.78      &  --       & 5.36      & --        & 3.0   & 13.01\% & --  & 18.68 \% \\
   & DCFlow+ft~\cite{XRK2017}                     & --        & 3.54      &  --       & 5.12      & --        & --    &   --  & -- & 14.83\% \\
   & FlowNet2+ft~\cite{ilg2017flownet}         & (1.45)    & 4.16      & (2.01)    & 5.74      & (1.28)    & 1.8   &  8.8\%& (2.3)  & 11.48\% \\
   & UnFlow-CSS+ft~\cite{Meister:2018:UUL}     &  --       &  --       &  --       & --        & (1.14)    & 1.7   &  8.42\%& (1.86) & 11.11\%\\
   & LiteFlowNet+ft-CVPR~\cite{hui18liteflownet} & (1.64)  & 4.86    & (2.23)    & 6.09      & (1.26)  & 1.7 &  -- &(2.16)&10.24\%\\
   & LiteFlowNet+ft-axXiv~\cite{hui18liteflownet}    & \textbf{(1.35)}& 4.54      & (1.78)    & 5.38      & (1.05)    & 1.6 &  7.27\%&(1.62)&9.38\%\\
   & PWC-Net+ft-CVPR~\cite{sun2018pwc}              &(2.02)     &4.39       & (2.08)    & 5.04      & (1.45)    & 1.7   & 8.10\%& (2.16) & 9.60\%  \\
   & PWC-Net+ft-axXiv~\cite{sun2018models}          & (1.71)    & 3.45     & (2.34) & 4.60 & (1.08) & \textbf{1.5} & 6.82\%  & (1.45) & 7.90\% \\
   & ProFlow+ft~\cite{Maurer_BMVC_2018_Proflow}&(1.78)     & \textbf{2.82} & --        & 5.02      & (1.89)    & 2.1   & 7.88\%  & (5.22) & 15.04\% \\
   & ContinualFlow+ft~\cite{Neoral2018ACCV}       & --        & 3.34      & --        & 4.52      & --        & --    & --    & --       & 10.03\% \\
   & MFF+ft~\cite{ren2018fusion}               & --        & 3.42      & --        & 4.57      & --        & 1.7   & 7.87\% & -- & \textbf{7.17\%} \\
   & Ours+ft                                   &(1.68)     & 3.74   &\textbf{(1.77)} &\textbf{4.26}& \textbf{(0.76)}&\textbf{1.5} &\textbf{6.19\%}&(\textbf{1.18}) & 8.42\%  \\
 \bottomrule \end{tabular} } 
\caption{Comparison with state-of-the-art learning based optical flow estimation methods. Our  method outperforms all unsupervised optical flow learning approaches on all datasets. Our supervised fine-tuned model achieves the highest accuracy on the Sintel Final dataset and KITTI 2012 dataset. All numbers are EPE except for the last column of  KITTI 2012 and KITTI 2015 testing sets, where we report percentage of erroneous pixels over all pixels (Fl-all). Missing entries (-) indicate that the results are not reported for the respective method. Parentheses mean that the training and testing are performed on the same dataset. Bold fonts highlight the best results among unsupervised and supervised methods respectively.
}
\vspace{-2ex}
\label{QuantitativeResult}
\end{table*}

\subsection{Occlusion  Estimation}
For two-frame optical flow estimation, we can swap two images as input to generate forward and backward flow, then the occlusion map can be generated based on the forward-backward consistency prior~\cite{sundaram2010dense,Meister:2018:UUL}. To make this work under our three-frame setting, we propose to utilize the adjacent five frame images as input as shown in Figure~\ref{OcclusionEstimation}. Specifically, we  estimate bi-directional flows between $I_t$ and $I_{t+1}$, namely $\textbf{w}_{t \to t+1}$ and $\textbf{w}_{t+1 \to t}$. Similarly, we  also estimate the flows between $I_t$ and $I_{t-1}$. Finally, we conduct a forward and backward consistency check  to reason the occlusion map between two consecutive images.

For forward-backward consistency check, we consider one pixel as occluded when the mismatch between the forward flow and the reversed forward flow is too large. Take $O_{t \to t+1}$ as an example, we can first compute the reversed forward flow as follows,
\begin{equation}
\hat{\textbf{w}}_{t \to t+1} = \textbf{w}_{t+1 \to t}({\textbf{p}+\textbf{w}_{t \to t+1}}(\textbf{p})),
\end{equation}
A pixel is considered occluded whenever it violates the following constraint:
\begin{equation}
|\textbf{w}_{t \to t+1} + \hat{\textbf{w}}_{t \to t+1}|^2 < \alpha_1 (|\textbf{w}_{t \to t+1}|^2+|\hat{\textbf{w}}_{t \to t+1}|^2) + \alpha_2,
\end{equation}
where we set $\alpha_1$ = 0.01, $\alpha_2$ = 0.05 for all our experiments. Other occlusion maps are computed in the same way.

\subsection{Occlusion Hallucination}
During our self-supervised training, we hallucinate occlusions by perturbing local regions with random noise. In a newly generated target image,  the pixels  corresponding to noise regions automatically become occluded.  There are many ways to generate such occlusions. The most straightforward way is to randomly select rectangle regions. However, rectangle occlusions rarely exist in real-world sequences. To address this issue, we propose to first generate superpixels~\cite{achanta2012slic}, then randomly select several superpixels and fill them with noise. There are two main advantages of using superpixel. First, the shape of a superpixel is usually random and superpixel edges are often part of object boundaries.  The is consistent with the real-world cases and makes the noise image more realistic. We can choose several superpixels which locate at different locations to cover more occlusion cases. Second, the pixels within each superpixel usually belong to the same object or have similar flow fields. Prior work has found low-level segmentation is helpful for  optical flow estimation~\cite{XRK2017}. Note that the random noise should lie in the pixel value range.

Figure~\ref{Framework} shows a simple example, where only the dog extracted from the COCO dataset~\cite{lin2014microsoft} is moving. Initially, the occlusion map between $I_t$ and $I_{t+1}$ is (g). After randomly selecting several superpixels from (e) to inject noise, the occlusion map between $I_t$ and $\widetilde{I}_{t+1}$ change to (h).
Next, we describe how to make use of these occlusion maps to guide our self-training.

\begin{figure*}[th]
 \centering
   \includegraphics[width=1\textwidth]{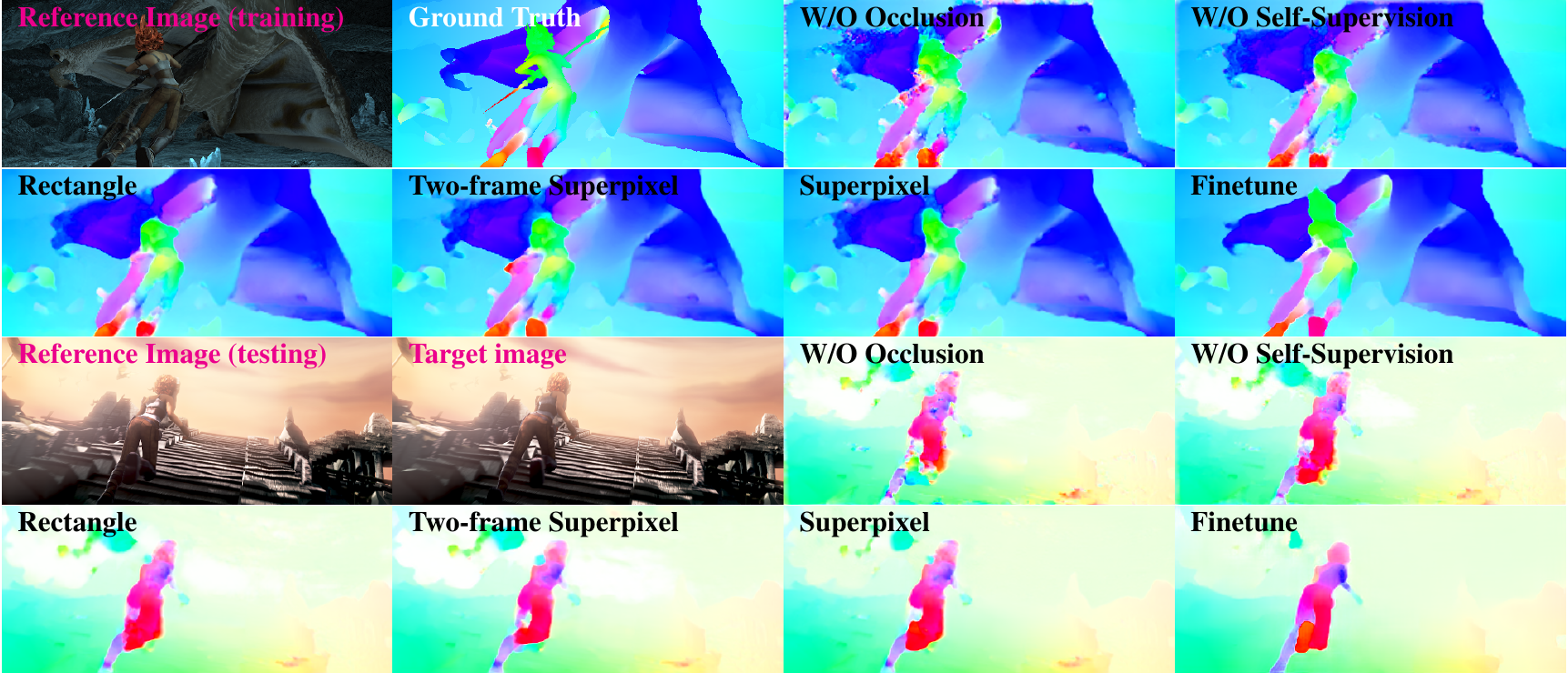}
 \caption{Qualitative comparison of our model under different settings on Sintel Clean training and Sintel Final testing dataset. Occlusion handling, multi-frame formulation and self-supervision consistently improve the performance.
}
\label{ComparisonSintel}
\end{figure*}

\subsection{NOC-to-OCC as Self-Supervision}
Our self-training idea is built on top of the classical  photometric loss~\cite{Meister:2018:UUL,wang2018occlusion,Janai2018ECCV}, which is highly effective for non-occluded pixels. Figure~\ref{Framework}  illustrates our main idea. Suppose  pixel $p_1$ in image $I_t$ is not occluded in $I_{t+1}$, and pixel $p'_{1}$ is its corresponding pixel. If we inject noise to $I_{t+1}$ and let $I_{t-1}$, $I_{t}$, $\widetilde{I}_{t+1}$ as input, $p_1$ then becomes occluded. Good news is we can still use the flow estimation of NOC-Model as annotations to guide OCC-Model to learn the flow of $p_1$ from $I_{t}$ to $\widetilde{I}_{t+1}$.
This is also consistent with  real-world occlusions, where the flow of occluded pixels can be estimated based on  surrounding non-occluded pixels. In the example of Figure~\ref{Framework}, self-supervision is only employed to (i), which represents those pixels non-occluded from $I_t$ to $I_{t+1}$ but become occluded from $I_t$ to $\widetilde{I}_{t+1}$.

\subsection{Loss Functions}
Similar to previous unsupervised methods, we first apply photometric loss $L_p$ to non-occluded pixels. Photometric loss is defined as follows:

\begin{equation}
   L_p = \sum_{i, j} \frac {\sum{\psi(I_i-I_{j \to i}^{w})} \odot (1 - O_i)}  {\sum{(1 - O_i)}}
  \label{eq:nocloss}
\end{equation}
where $\psi(x) = (|x|+\epsilon)^q$ is a robust loss function, $\odot$ denotes the element-wise multiplication. We set $\epsilon=0.01$, $q=0.4$ for all our experiments. Only $L_p$ is necessary to train the NOC-Model.

To train our  OCC-Model to estimate  optical flow of occluded pixels, we define a self-supervision loss
$L_o$ for those synthetic occluded pixels (Figure~\ref{Framework}(i)).
First, we compute a self-supervision mask $M$ to represent these pixels,
\begin{equation}
M_{i \to j} = \text{clip}(\widetilde{O}_{i \to j}-O_{i \to j}, 0, 1)
\end{equation}

Then, we  define our self-supervision loss $L_o$ as,
\begin{equation}
   L_o = \sum_{i, j} \frac {\sum{\psi(\textbf{w}_{i \to j}-\widetilde{\textbf{w}}_{i \to j}) \odot M_{i \to j}}}  {\sum{M_{i \to j}}}
  \label{eq:occloss}
\end{equation}
For our OCC-Model, we train with a simple combination of $L_p+L_o$  for both non-occluded pixels and occluded pixels.
Note our loss functions do not rely on spatial and temporal consistent assumptions, and they can be used for both  classical two-frame flow estimation and multi-frame flow estimation.

\subsection{Supervised Fine-tuning}
After pre-training on raw dataset, we use real-world annotated data for fine-tuning. Since there are only annotations for forward flow $\textbf{w}_{t \to t+1}$, we skip backward flow estimation when computing our loss. Suppose that the ground truth flow is $\textbf{w}^{gt}_{t \to t+1}$, and mask $V$ denotes whether the pixel has a label, where value 1 means that the pixel has a valid ground truth flow.  Then we can obtain the supervised fine-tuning loss    as follows,
\begin{equation}
L_s = \sum(\psi(\textbf{w}^{gt}_{t \to t+1} - \textbf{w}_{t \to t+1}) \odot V) / \sum{V}
\end{equation}

During fine-tuning, We first initialize the model with the pre-trained OCC-Model on each dataset, then optimize it using $L_s$.

\section{Experiments}
We evaluate and compare our methods with state-of-the-art unsupervised and supervised learning methods on public optical flow benchmarks including MPI Sintel~\cite{butler2012naturalistic},  KITTI 2012~\cite{geiger2012we} and KITTI 2015~\cite{menze2015object}.
To ensure reproducibility and advance further innovations, we  make our code and models publicly available
at \texttt{https://github.com/ppliuboy/SelFlow}.

\begin{figure*}[t]
 \centering
   \includegraphics[width=1\textwidth]{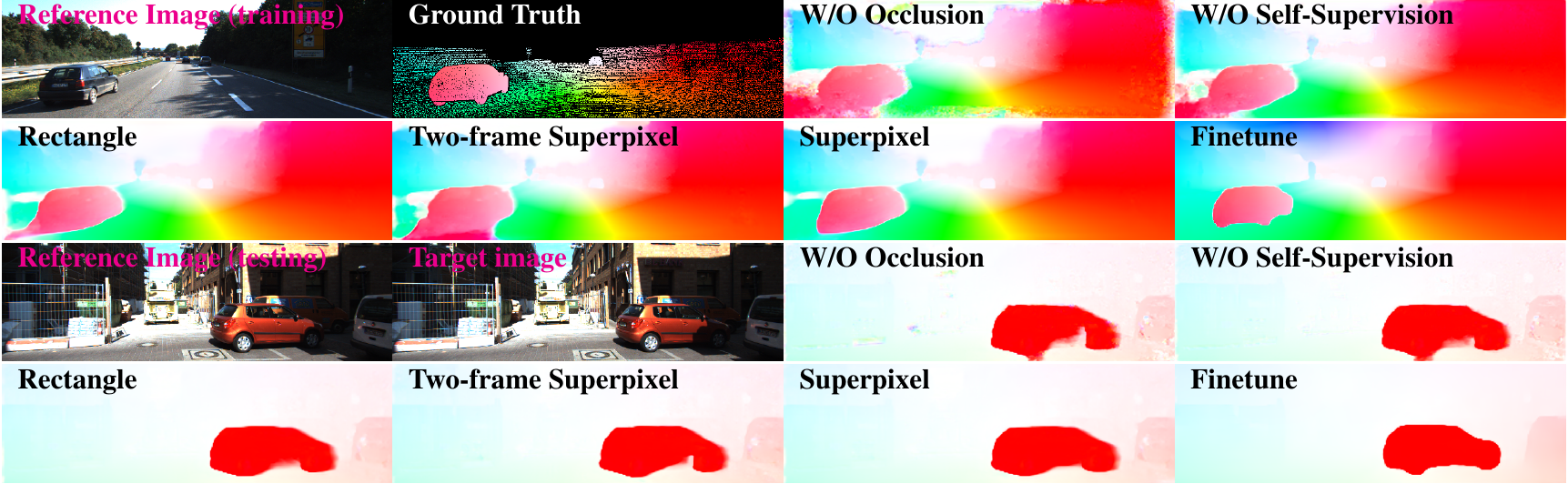}
 \caption{Qualitative comparison of our model under different settings on KITTI 2015 training and testing dataset. Occlusion handling, multi-frame formulation and self-supervision consistently improve the performance.
}
\label{ComparisonKitti}
\end{figure*}

\subsection{Implementation Details}

\mypara{Data Preprocessing.}
For Sintel, we download the Sintel movie and extract  $\sim10,000$ images for self-training. We first train our model on this raw data, then add the official Sintel training data (including both "final" and "clean" versions). For KITTI 2012 and KITTI 2015, we use multi-view extensions of the two datasets for unsupervised pre-training, similar to~\cite{ren2017unsupervised,wang2018occlusion}. During training, we exclude the image pairs with ground truth flow and their neighboring frames (frame number 9-12) to avoid the mixture of training and testing data.

We rescale the pixel value from [0, 255] to [0, 1] for unsupervised training, while normalizing each channel to be standard normal distribution for supervised fine-tuning. This is because normalizing image as input is more robust for luminance changing, which is especially helpful for optical flow estimation. For unsupervised training,  we apply Census Transform~\cite{zabih1994non} to images, which has been proved robust for optical flow estimation~\cite{hafner2013census,Meister:2018:UUL}.

\mypara{Training procedure.}
We train our model with the Adam optimizer~\cite{kingma2014adam} and set batch size to be 4 for all experiments. For unsupervised training, we set the initial learning rate to be $10^{-4}$, decay it by half every 50k iterations, and use random cropping, random flipping, random channel swapping during data augmentation.
For supervised fine-tuning, we employ similar data augmentation and learning rate schedule as~\cite{dosovitskiy2015flownet,ilg2017flownet}.

For unsupervised pre-training, we first train our NOC-Model with photometric loss for 200k iterations. Then, we add our occlusion regularization and train for another 500k iterations. Finally, we initialize the OCC-Model with the trained weights of NOC-Model and train it with $L_p+L_o$ for 500k iterations. Since training two models simultaneously will cost more memory and  training time, we just generate the flow and occlusion maps using the NOC-Model in advance and use them as annotations (just like KITTI with sparse annotations).

For supervised fine-tuning, we use the pre-trained OCC-Model as initialization, and train the model using our supervised loss $L_s$ with $500$k iterations for KITTI and $1,000$k iterations for Sintel. Note we do not require pre-training our model on any labeled synthetic dataset, hence we  do not have to follow the specific training schedule (FlyingChairs~\cite{dosovitskiy2015flownet}$\to$ FlyingThings3D~\cite{mayer2016large}) as~\cite{ilg2017flownet,hui18liteflownet,sun2018pwc}.

\mypara{Evaluation Metrics.}
We consider two widely-used metrics to evaluate optical flow estimation: average endpoint error (EPE), percentage of erroneous pixels (Fl). EPE is the ranking metric on the Sintel benchmark, and Fl is the ranking metric on KITTI benchmarks.

  \begin{table*}[t]
  \centering
  \resizebox{\textwidth}{!}{
  \begin{tabular}{ c c c    c c c  c c c  c c c  c cc c c c}
   \toprule
     Occlusion   &Multiple & Self-Supervision & Self-Supervision &\multicolumn{3}{c}{Sintel Clean}& \multicolumn{3}{c}{Sintel Final} & \multicolumn{3}{c}{KITTI 2012} & \multicolumn{3}{c}{KITTI 2015} \\
    \cmidrule(l{3mm}r{3mm}){5-7}   \cmidrule(l{3mm}r{3mm}){8-10}   \cmidrule(l{3mm}r{3mm}){11-13}   \cmidrule(l{3mm}r{3mm}){14-16}
     Handling    &Frame &  Rectangle  & Superpixel& ALL  & NOC   &  OCC   &  ALL  &  NOC  &  OCC   &   ALL &  NOC  &  OCC  &   ALL  &   NOC   &   OCC \\
    \midrule

  \xmark &\xmark & \xmark & \xmark & (3.85) & (1.53) & (33.48) & (5.28) & (2.81) & (36.83) & 7.05 & 1.31 & 45.03 & 13.51 & 3.71 & 75.51 \\
  \xmark &\cmark & \xmark & \xmark & (3.67) & (1.54) & (30.80) & (4.98) & (2.68) & (34.42) & 6.52 & 1.11 & 42.44 & 12.13 & 3.47 & 66.91 \\
  \cmark &\xmark & \xmark & \xmark & (3.35) & (1.37) & (28.70) & (4.50) & (2.37) & (31.81) & 4.96 & 0.99 & 31.29 & 8.99  & 3.20 & 45.68 \\
  \cmark &\cmark & \xmark & \xmark & (3.20) & (1.35) & (26.63) & (4.33) & (2.32) & (29.80) & 3.32 & 0.94 & 19.11 & 7.66  & 2.47 & 40.99 \\
  \cmark &\xmark & \xmark & \cmark & (2.96) & (1.33) & (23.78) & (4.06) & (2.25) & (27.19) & 1.97 & 0.92 & 8.96 & 5.85 & 2.96 & 24.17 \\
  \cmark &\cmark & \cmark & \xmark & (2.91) & (1.37) & (22.58) & (3.99) & (2.27) & (26.01) & 1.78 & 0.96 & 7.47  & 5.01  & 2.55 & 21.86 \\
  \cmark &\cmark & \xmark & \cmark & \textbf{(2.88)} & \textbf{(1.30)} & \textbf{(22.06)} & \textbf{(3.87)} & \textbf{(2.24)} & \textbf{(25.42)} & \textbf{1.69} & \textbf{0.91} & \textbf{6.95} & \textbf{4.84} & \textbf{2.40} & \textbf{19.68} \\
   \bottomrule
  \end{tabular} }
  \caption{Ablation study. We report EPE of our unsupervised results under different settings over all pixels (ALL), non-occluded pixels (NOC) and occluded pixels (OCC). Note that we employ Census Transform when computing photometric loss by default. Without Census Transform, the performance will drop.}
  \label{AblationStudy}
  \end{table*}

  \begin{table}[th]
  \centering
  \resizebox{0.5\textwidth}{!}{
  \begin{tabular}{ c c c c  c }
   \toprule
     Unsupervised Pre-training  &  Sintel Clean & Sintel Final  &  KITTI 2012 & KITTI 2015    \\
    \midrule
     Without                    &     1.97      &      2.68     &    3.93     &      3.10     \\
       With                     & \textbf{1.50} & \textbf{2.41} &\textbf{1.55}& \textbf{1.86} \\
   \bottomrule
  \end{tabular} } 
  \caption{Ablation study. We report EPE of supervised fine-tuning results on our validation datasets with and without unsupervised pre-training.}
  \label{AblationPretrainSupervise}
  \end{table}

\subsection{Main Results}
As shown in Table~\ref{QuantitativeResult}, we achieve  state-of-the-art results for both unsupervised and supervised optical flow learning  on all datasets under all evaluation metrics.
Figure~\ref{QualitativeResult} shows sample results from Sintel and KITTI. Our method estimates both accurate optical flow and occlusion maps.

\mypara{Unsupervised Learning.}  Our method achieves the highest accuracy for unsupervised learning methods on leading benchmarks.
On the Sintel final  benchmark, we reduce the previous best EPE from $7.40$~\cite{Liu:2019:DDFlow} to $6.57$, with $11.2\%$ relative improvements. This is even better than several fully supervised methods including FlowNetS, FlowNetC~\cite{dosovitskiy2015flownet}, and SpyNet~\cite{ranjan2017optical}.

On the KITTI datasets, the improvement is more significant. For the training dataset, we achieve EPE=1.69 with $28.1\%$ relative improvement on KITTI 2012 and EPE=$4.84$ with $15.3\%$ relative improvement on KITTI 2015 compared with previous best unsupervised method DDFlow. On KITTI 2012  testing set, we achieve Fl-all=$7.68\%$, which is better than state-of-the-art supervised methods including FlowNet2~\cite{ilg2017flownet}, PWC-Net~\cite{sun2018pwc}, ProFlow~\cite{Maurer_BMVC_2018_Proflow}, and MFF~\cite{ren2018fusion}. On KITTI 2015 testing benchmark, we achieve Fl-all 14.19\%, better than all unsupervised methods. Our unsupervised results also outperform some fully supervised methods including DCFlow~\cite{XRK2017}  and ProFlow~\cite{Maurer_BMVC_2018_Proflow}.

\mypara{Supervised Fine-tuning.} We further fine-tune our unsupervised model with the ground truth flow. We achieve state-of-the-art results on all three datasets, with Fl-all=6.19\% on KITTI 2012 and Fl-all=8.42\% on KITTI 2015. Most importantly, our method yields EPE=4.26 on the Sintel final dataset, achieving the highest accuracy on the  Sintel benchmark among all submitted methods. All these show that our method reduces the reliance of pre-training with synthetic datasets and we do not have to follow specific training schedules across different datasets anymore.

\subsection{Ablation Study}
To  demonstrate the usefulness of individual technical steps, we conduct a rigorous ablation study  and show the quantitative comparison  in Table~\ref{AblationStudy}. Figure~\ref{ComparisonSintel} and Figure~\ref{ComparisonKitti} show the qualitative comparison under different settings, where ``W/O Occlusion'' means occlusion handling is not considered, ``W/O Self-Supervision'' means occlusion handling is considered but self-supervision is not employed, ``Rectangle'' and ``Superpixel'' represent self-supervision is employed with rectangle and superpixel noise injection respectively. ``Two-Frame Superpixel'' means self-supervision is conducted with only two frames as input.

\mypara{Two-Frame vs Multi-Frame.}
Comparing row 1 and row 2, row 3 and row 4 row 5 and row 7 in Table~\ref{AblationStudy}, we can see that using multiple frames as input can indeed improve the performance, especially for occluded pixels. It is because multiple images provide more information, especially for those pixels occluded in one direction but non-occluded in the reverse direction.

\mypara{Occlusion Handling.}
Comparing the row 1 and row 3, row 2 and row 4 in Table~\ref{AblationStudy}, we can see that occlusion handling can improve  optical flow estimation performance over all pixels on all datasets.  This is due to the fact that brightness constancy assumption does not hold for occluded pixels.

\mypara{Self-Supervision.}
We employ two strategies for our occlusion hallucination: rectangle and superpixel. Both strategies improve the performance significantly, especially for occluded pixels. Take superpixel setting as an example, EPE-OCC decrease from 26.63 to 22.06 on Sintel Clean, from 29.80 to 25.42  on Sintel Final, from 19.11 to 6.95  on KITTI 2012, and from 40.99 to 19.68  on KITTI 2015. Such a big improvement demonstrates the effectiveness of our self-supervision strategy.

Comparing superpixel noise injection with rectangle noise injection, superpixel setting has several advantages. First, the shape of the superpixel is random and edges are more correlated to motion boundaries. Second, the pixels in the same superpixel usually have similar motion patterns. As a result, the superpixel setting achieves slightly better performance.

\mypara{Self-Supervised  Pre-training.} Table~\ref{AblationPretrainSupervise} compares supervised results with and without our self-supervised  pre-training on the validation sets. If we do not employ self-supervised  pre-training and directly train the model using only the ground truth, the model fails to converge well due to insufficient training data. However, after utilizing our self-supervised  pre-training, it converges very quickly and achieves much better results.

\section{Conclusion}
We have presented  a self-supervised approach to learning accurate optical flow estimation. Our method injects noise into superpixels to create occlusions, and let one model guide the another   to learn  optical flow for  occluded pixels. Our simple CNN effectively aggregates temporal information from multiple frames   to improve flow prediction. Extensive experiments show our method significantly outperforms all existing unsupervised optical flow learning methods. After fine-tuning with our unsupervised model, our method  achieves state-of-the-art flow estimation accuracy on all leading benchmarks. Our results demonstrate it is possible to completely reduce the reliance of pre-training on synthetic labeled datasets, and achieve superior performance by self-supervised pre-training on unlabeled data.

\section{Acknowledgment}
This work is supported by the Research Grants Council of the Hong Kong Special Administrative Region, China (No. CUHK 14208815 and No. CUHK 14210717 of the General Research Fund).  We thank anonymous reviewers for their constructive suggestions. 

\balance

{\small
\bibliographystyle{ieee_fullname}
\bibliography{egbib}
}

\clearpage
\twocolumn[
  \begin{@twocolumnfalse}
{
   \newpage
   \null
   \vskip .375in
   \begin{center}
      {\Large \bf Supplementary Material \par}
      \vspace*{24pt}
      {
      \large
      \lineskip .5em
      \begin{tabular}[t]{c}
          
      \end{tabular}
      \par
      }
      \vskip .5em
      \vspace*{12pt}
   \end{center}
}
  \end{@twocolumnfalse}
]
\setcounter{section}{0}
\setcounter{figure}{0}
\setcounter{table}{0}
\setcounter{footnote}{0}
\renewcommand*{\theHsection}{A\thesection}
\renewcommand*{\theHfigure}{A\thefigure}
\renewcommand*{\theHtable}{A\thetable}
\section{Overview}
In this supplement, we first show occlusion estimation performance of SelFlow. Then we present screenshots (Nov. 23, 2018) of our submission on the public benchmarks, including MPI Sintel final pass, KITTI 2012, and KITTI 2015. 

\section{Occlusion Estimation}
Following~\cite{wang2018occlusion,Janai2018ECCV,Liu:2019:DDFlow}, we also report the occlusion estimation performance using F-measure, which is the harmonic mean of precision and recall. We estimate occlusion map using forward-backward consistency check (no parameters to learn). 

We compare our occlusion estimation performance with MODOF~\cite{xu2012motion}, OccAwareFlow~\cite{wang2018occlusion}, MultiFrameOccFlow-Soft~\cite{Janai2018ECCV} and DDFlow. Note KITTI datasets only have sparse occlusion maps. As shown in Table~\ref{OcclusionComparison}, we achieve the best occlusion estimation performance on Sintel Clean and Sintel Final, and comparable performance on KITTI 2012 and 2015.

\begin{table}[!htbp]   \centering
\resizebox{0.48\textwidth}{!}{
\begin{tabular}{ l c c c c }
 \toprule   \multirow{2}{*}{Method}                      & Sintel  & Sintel  & KITTI    & KITTI \\
                               & Clean   & Final   & 2012     & 2015  \\
  \midrule
   MODOF                         &   --    & 0.48    &  --      & --   \\
   OccAwareFlow               &  (0.54) & (0.48)  & \textbf{0.95$^*$} & 0.88$^*$   \\
   MultiFrameOccFlow-Soft    &  (0.49) & (0.44)  &  --      & \textbf{0.91$^*$} \\
   DDFlow                     & \textbf{(0.59)} &\textbf{(0.52)} & 0.94 $^*$ & 0.86 $^*$ \\
   Ours                          & \textbf{(0.59)} & \textbf{(0.52)}  &\textbf{0.95} $^*$ & 0.88$^*$ \\
  \bottomrule \end{tabular}}
\caption{Comparison of occlusion estimation with F-measure.  $^*$ marks cases where the occlusion annotation is sparse.
} \label{OcclusionComparison}
\end{table}

\section{Screenshots on Benchmarks}
\begin{figure*}[t]
 \centering
   \includegraphics[width=1\textwidth]{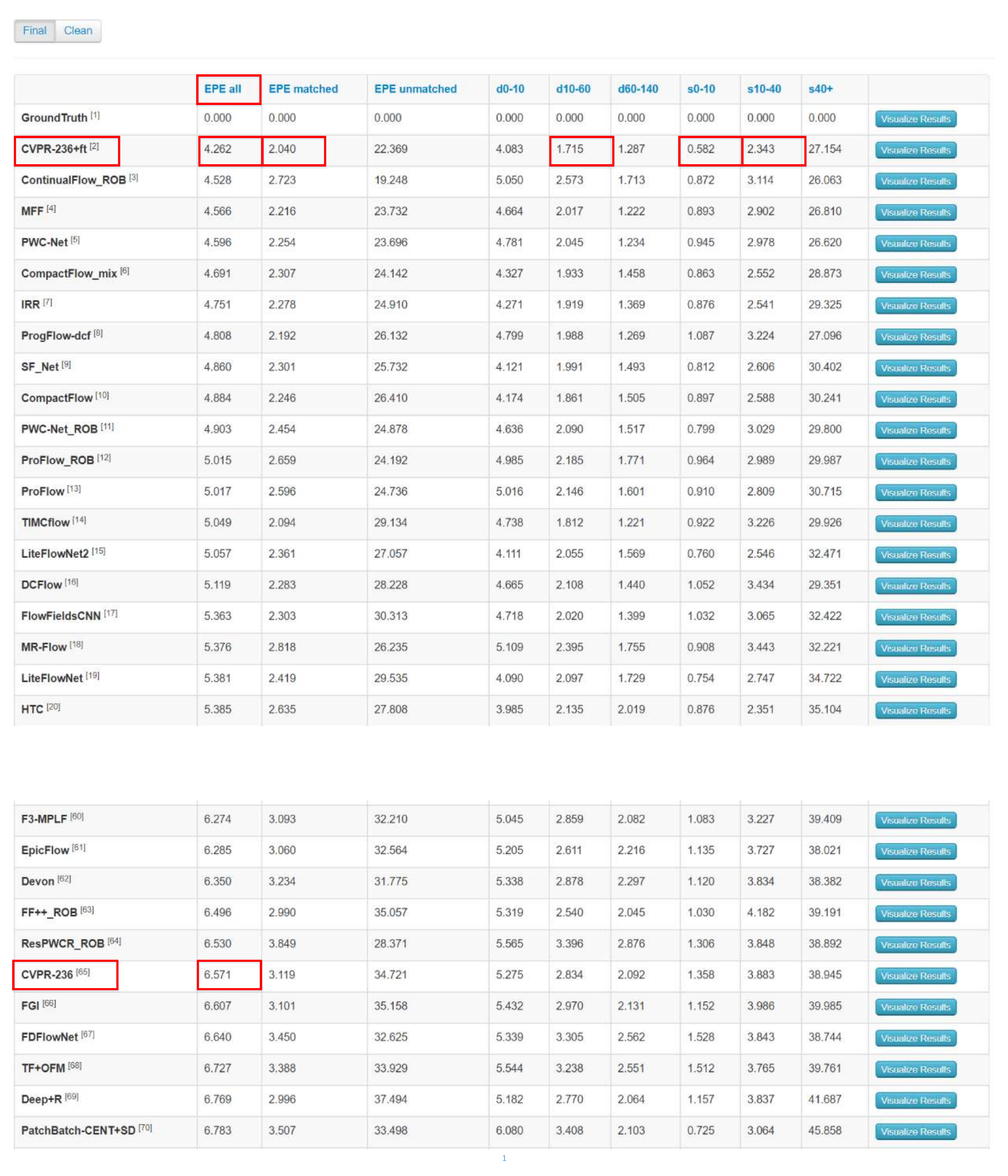}
   \vspace{-2mm}
 \caption{Screenshot of the Sintel benchmark on November 23th, 2018.
}
 \label{ScreenshotSintel}
 \end{figure*}

\begin{figure*}[t]
 \centering
   \includegraphics[width=1\textwidth]{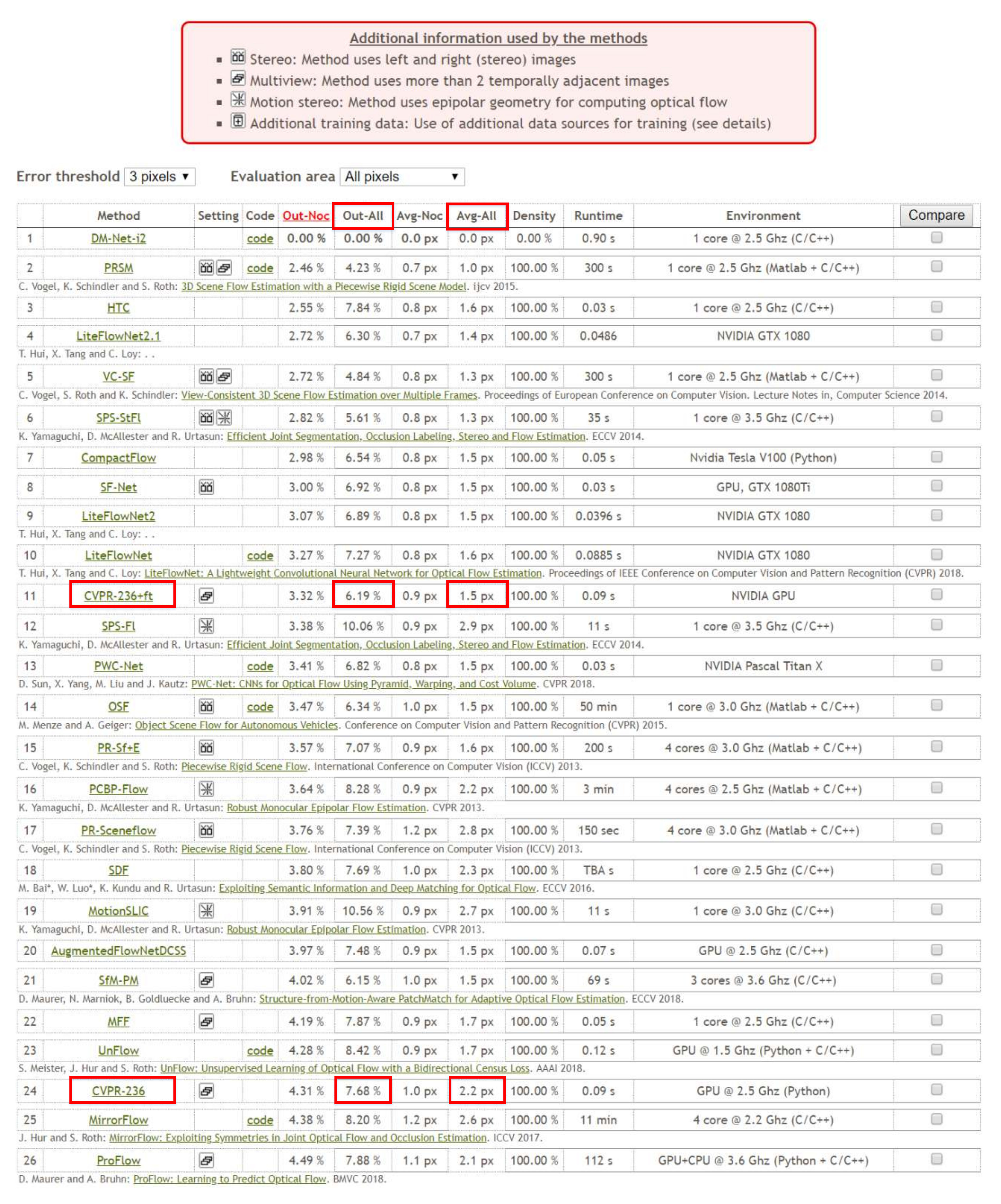}
   \vspace{-2mm}
 \caption{Screenshot of the KITTI 2012 benchmark on November 23th, 2018.
}
 \label{ScreenshotKitti2012}
 \end{figure*}

\begin{figure*}[t]
 \centering
   \includegraphics[width=1\textwidth]{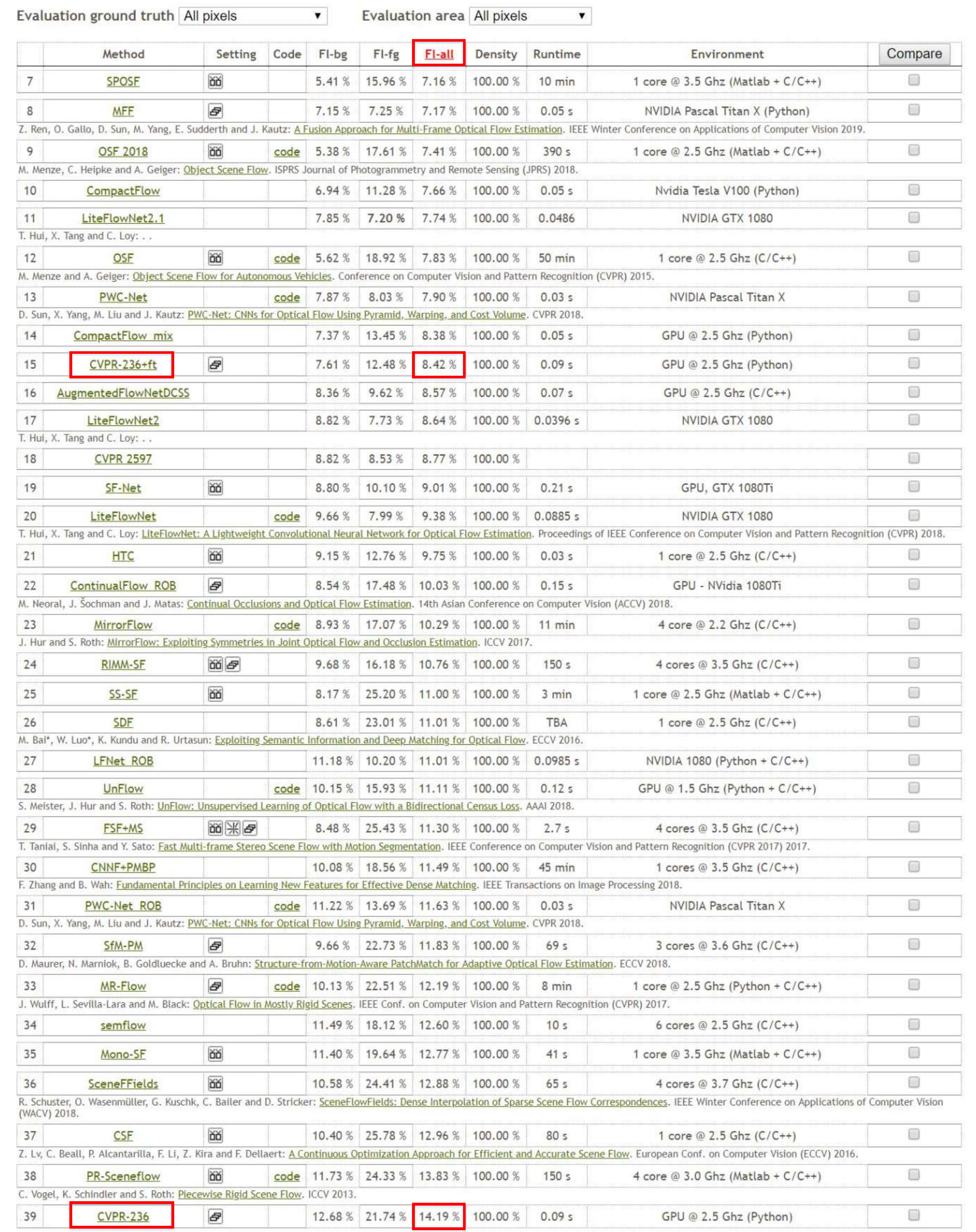}
   \vspace{-4.5mm}
 \caption{Screenshot of the KITTI 2015 benchmark on November 23th, 2018.
}
 \label{ScreenshotKitti2015}
 \end{figure*}

Figure~\ref{ScreenshotSintel} shows the screenshot of our submission on the MPI Sintel benchmark. Our unsupervised entry (CVPR-236) outperforms all the exiting unsupervised learning method, even outperforming supervised methods including FlowNetS+ft+v, FlowNetC+ft+v and SpyNet+ft. At the time of writing, our supervised fine-tuned entry (CVPR-236+ft) is the No. 1 among all submitted methods. In addition to the main ranking metric EPE-all, our method also achieves the best performance on
EPE-matched, d10-60, s0-10, s10-40, and very competitive results on remaining metrics. This clearly demonstrates the
effectiveness of our method.
Figure~\ref{ScreenshotKitti2012} and Figure~\ref{ScreenshotKitti2015} show the screenshots of KITTI 2012 and KITTI 2015 benchmark. Again, our unsupervised entry (CVPR-236) outperforms all the exiting unsupervised learning method on both benchmarks. On KITTI 2012, our unsupervised entry (CVPR-236) even outperforms the most recent fully supervised methods including ProFlow, ImpPB+SPCI, Flow-FieldCNN, IntrpNt-df. Our supervised fine-tuned entry (CVPR-236+ft) is the second best compared to published monocular optical flow estimation methods (only second to LiteFlowNet), while achieving better Out-All and Ave-All. On KITTI 2015, our unsupervised entry (CVPR-236) also outperforms several recent supervised methods including DCFlow, ProFlow, FlowFields++ and FlowFieldCNN. Our supervised fine-tuned entry (CVPR-236+ft) is the third best compared to published monocular optical flow estimation methods, only behind the concurrent work MFF, and the extended version of PWC-Net.

\end{document}